\documentclass[twocolumn,a4paper,10pt]{svjour3}
\journalname{IJCV}

\usepackage{boilerplate}
\usepackage[utf8]{inputenc}
\usepackage[round]{natbib}

\renewcommand{\a}{\mathbf{a}}
\newcommand{\x}{\mathbf{x}}
\newcommand{\z}{\mathbf{z}}

\newcommand{\y}{\mathbf{y}}

\newcommand{\s}{\mathbf{s}}
\renewcommand{\d}{\mathbf{d}}

\renewcommand{\l}{\boldsymbol{\lambda}}

\let\secsym\S
\newcommand{\fig}[1]{Fig. \ref{fig:#1}}
\renewcommand{\sec}[1]{\secsym\ref{sec:#1}}
\newcommand{\eqn}[1]{Eqn. \ref{eqn:#1}}
\newcommand{\app}[1]{Appendix \ref{sec:#1}}

\newcommand{\df}{\hat{\mathbf{d}}}
\newcommand{\xf}{\hat{\mathbf{x}}}
\renewcommand{\sf}{\hat{\mathbf{s}}}
\newcommand{\zf}{\hat{\mathbf{z}}}
\renewcommand{\df}{\hat{\mathbf{d}}}
\newcommand{\tf}{\hat{\mathbf{t}}}
\renewcommand{\zf}{\hat{\mathbf{z}}}
\newcommand{\lf}{\hat{\l}}
\newcommand{\Df}{\hat{\mathbf{D}}}
\newcommand{\Zf}{\hat{\mathbf{Z}}}

\renewcommand{\t}{\mathbf{t}}
\renewcommand{\u}{\mathbf{u}}

\renewcommand{\c}{\mathbf{c}}

\newcommand{\R}{\mathbb{R}}

\newcommand{\B}{\mathbf{B}}

\renewcommand{\L}{\mathcal{L}}

\newcommand{\F}{\mathbf{F}}
\newcommand{\I}{\mathbf{I}}
\newcommand{\A}{\mathbf{A}}
\newcommand{\D}{\mathbf{D}}
\renewcommand{\O}{\mathcal{O}}

\newcommand{\Q}{\mathbf{Q}}

\renewcommand{\S}{\mathbf{S}}
\newcommand{\X}{\mathbf{X}}
\newcommand{\Z}{\mathbf{Z}}
\newcommand{\Phib}{\boldsymbol{\Phi}}

\newcommand{\dft}{\mathcal{F}}

\newcommand{\diag}{\mbox{diag}}

\newcommand{\conv}{\ast}
\newcommand{\corr}{\star}

\newcommand{\sgn}{\mbox{sgn}}

\newcommand{\st}{\mbox{s.t.}}
\newcommand{\grad}{\nabla}
\newcommand{\prox}{\mbox{prox}}

\newcommand{\half}{\frac{1}{2}}

\makeatletter
\DeclareRobustCommand\onedot{\futurelet\@let@token\@onedot}
\def\@onedot{\ifx\@let@token.\else.\null\fi\xspace}

\def\eg{\emph{e.g}\onedot} 
\def\ie{\emph{i.e}\onedot} 
 
\def\etc{\emph{etc}\onedot} 
 
\def\etal{\emph{et al}\onedot}
\makeatother

\title{Optimization Methods for Convolutional Sparse Coding}
\author{Hilton Bristow \and Simon Lucey}
\institute{Hilton Bristow 
	\at The Queensland University of Technology\\Brisbane, Australia \\
	\email{\href{mailto:hilton.bristow+papers@gmail.com}{hilton.bristow@gmail.com}}
\and Simon Lucey 
	\at Carnegie Mellon University\\ Pittsburgh, USA \\
	\email{\href{mailto:slucey@cs.cmu.edu}{slucey@cs.cmu.edu}}
}

\date{\today}

\begin{document}
\maketitle

\begin{abstract}
Sparse and convolutional constraints form a natural prior for many optimization problems that arise from physical processes. Detecting motifs in speech and musical passages, super-resolving images, compressing videos, and reconstructing harmonic motions can all leverage redundancies introduced by convolution. Solving problems involving sparse and convolutional constraints remains a difficult computational problem, however. In this paper we present an overview of \emph{\mbox{convolutional sparse coding}} in a consistent framework. The objective involves iteratively optimizing a convolutional least-squares term for the basis functions, followed by an $L_1$-regularized least squares term for the sparse coefficients. We discuss a range of optimization methods for solving the convolutional sparse coding objective, and the properties that make each method suitable for different applications. In particular, we concentrate on computational complexity, speed to $\epsilon$ convergence, memory usage, and the effect of implied boundary conditions. We present a broad suite of examples covering different signal and application domains to illustrate the general applicability of convolutional sparse coding, and the efficacy of the available optimization methods.

\end{abstract}

\section{Introduction}
Convolutional constraints arise in response to many natural phenomena. They provide a means of expressing an intuition that the relationships between variables should be both spatially or temporally localized, and redundant across the support of the signal. 

Coupled with sparsity, one can also enforce selectivity, \ie that the patterns observed in the signal stem from some structured underlying basis whose phase alignment is arbitrary.

In this paper, we discuss a number of optimization methods designed to solve problems involving convolutional constraints and sparse regularization in the form of the convolutional sparse coding algorithm,\footnote{Convolutional Sparse Coding has also been coined \emph{Shift Invariant} Sparse Coding (SISC), however the authors believe the term ``convolutional'' is more representative of the algorithm's properties.} and show through examples that the algorithm is well suited to a wide variety of problems that arise in computer vision.

\begin{figure}
\includegraphics[width=\columnwidth]{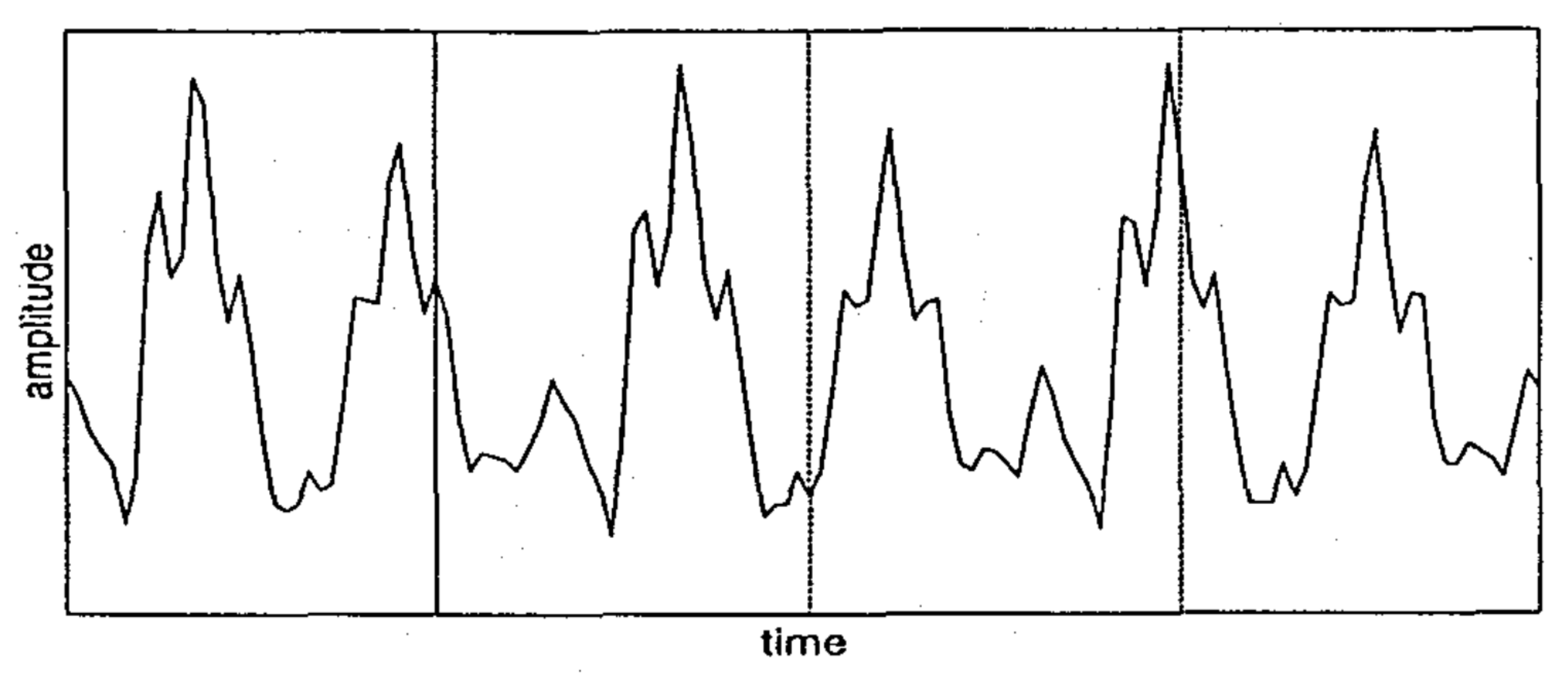}
\caption{Blocking results in arbitrary alignment of the underlying signal structure, artificially inflating the rank of the basis required to reconstruct the signal. Convolutional sparse coding relaxes this constraint, allowing shiftable basis functions to discover a lower rank structure.
\label{fig:blocking}}
\end{figure}
Consider the signal in \fig{blocking} from~\citep{lewicki_NIPS_1999}. The signal is composed of two distinct modes, appearing at multiple intervals (you could consider the modes to be expressions that are repeated over the course of a conversation, features that appear multiple times within an image, or particular motifs in a musical passage or speech). We wish to discover these modes and their occurrences in the signal in an unsupervised manner.

If the signal is segmented into blocks, the latent structure becomes obfuscated, and the basis learned must have the capacity to reconstruct each block in isolation. In effect, we learn a basis that is higher rank than the true basis due to the artificial constraints placed on the temporal alignment of the bases.

Convolutional sparse coding makes no such assumptions, allowing shiftable basis functions to discover a lower rank structure, and should therefore be preferred in situations where the basis alignment is not known \emph{a priori}. As we aim to demonstrate in this paper, this arises in a large number of practical applications.




\begin{figure*}
\includegraphics[width=\textwidth]{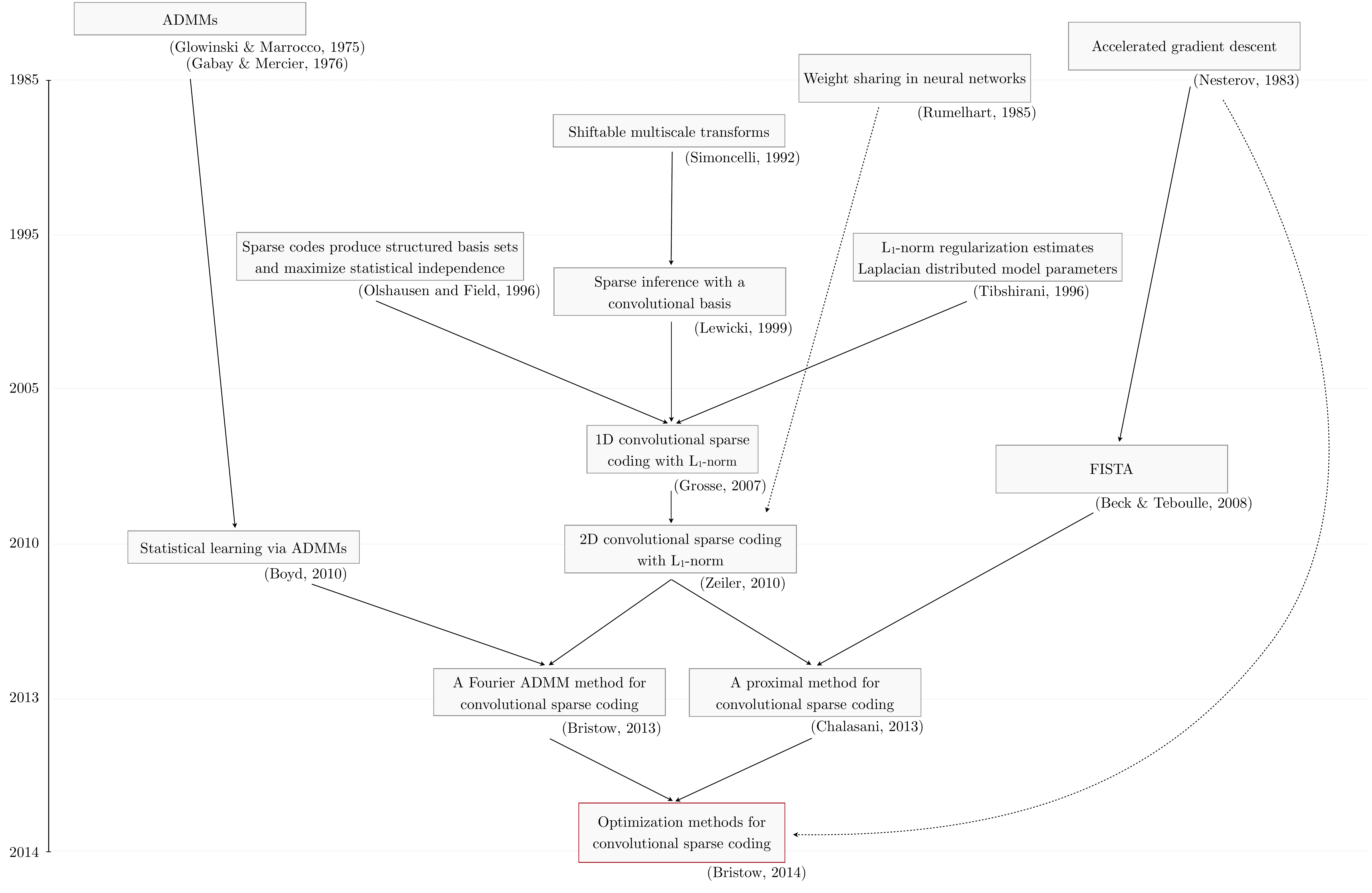}
\caption{A brief history of the works that influenced the direction of the convolutional sparse coding problem, and how it is optimized. The text within each box indicates the theme or idea that the paper introduced.
\label{fig:chronology}}
\end{figure*}

The notion of translation invariant optimization stems from the thought experiments of Simoncelli~\etal~\citep{simoncelli_IT_1992}. His motivation for considering translation invariance came from the context of wavelet transforms for signal processing. Amongst others, he had observed that block-based wavelet algorithms were sensitive to translation and scaling of the input signal. 

As an example, he chose an input signal to be one of the wavelet basis functions (yielding a single reconstruction coefficient), then perturbed that signal slightly to produce a completely dense set of coefficients. The abrupt change in representation in the wavelet domain due to a small change in the input illustrated the wavelet transform's unsuitability for higher-level summarization.


Olshausen and Field showed that sparsity alone is a sufficient driver for learning structured overcomplete representations from signals, and used it to learn a basis for natural image patches~\citep{olshausen_field_NATURE_1996}. The resulting basis, featuring edges at different scales and orientations, was similar to the receptive fields observed in the primary visual cortex.

The strategy of sampling patches from natural images has come under fire however, since many of the learned basis elements are simple translations of each other - an artefact of having to reconstruct individual patches, rather than entire image scenes~\citep{kavukcuoglu_NIPS_2010}. Removing the artificial assumption that image patches are independent - by modelling interactions in a convolutional objective - results in more expressive basis elements that better explain the underlying mechanics of the signal.

Lewicki and Sejnowski~\citep{lewicki_NIPS_1999} made the first steps towards this realization, by finding a set of sparse coefficients (value and temporal position) that reconstructed the signal with a fixed basis. They remarked at the spike-like responses observed, and the small number of coefficients needed to achieve satisfactory reconstructions.

Introducing sparsity brought with it a set of computational challenges that made the resulting objectives difficult to optimize. \citep{olshausen_field_NATURE_1996} explored coefficients drawn from a Cauchy distribution (a smooth heavy-tailed distribution) and a Laplacian distribution (a non-smooth heavy-tailed distribution), citing that in both cases \emph{they favour among activity states with equal variance, those with fewest non-zero coefficients}. \citep{olshausen_field_NATURE_1996} inferred the coefficients as the equilibrium solution to a differential equation. \citep{lewicki_NIPS_1999} assumed Laplacian distributed coefficients, and noted that due to the high sparsity of the desired response, it would be sufficient to replace exact inference with a procedure for guessing the values and temporal locations of the non-zero coefficients, then refining the results through a modified conjugate gradient local search.

\begin{figure}
\includegraphics[width=\columnwidth]{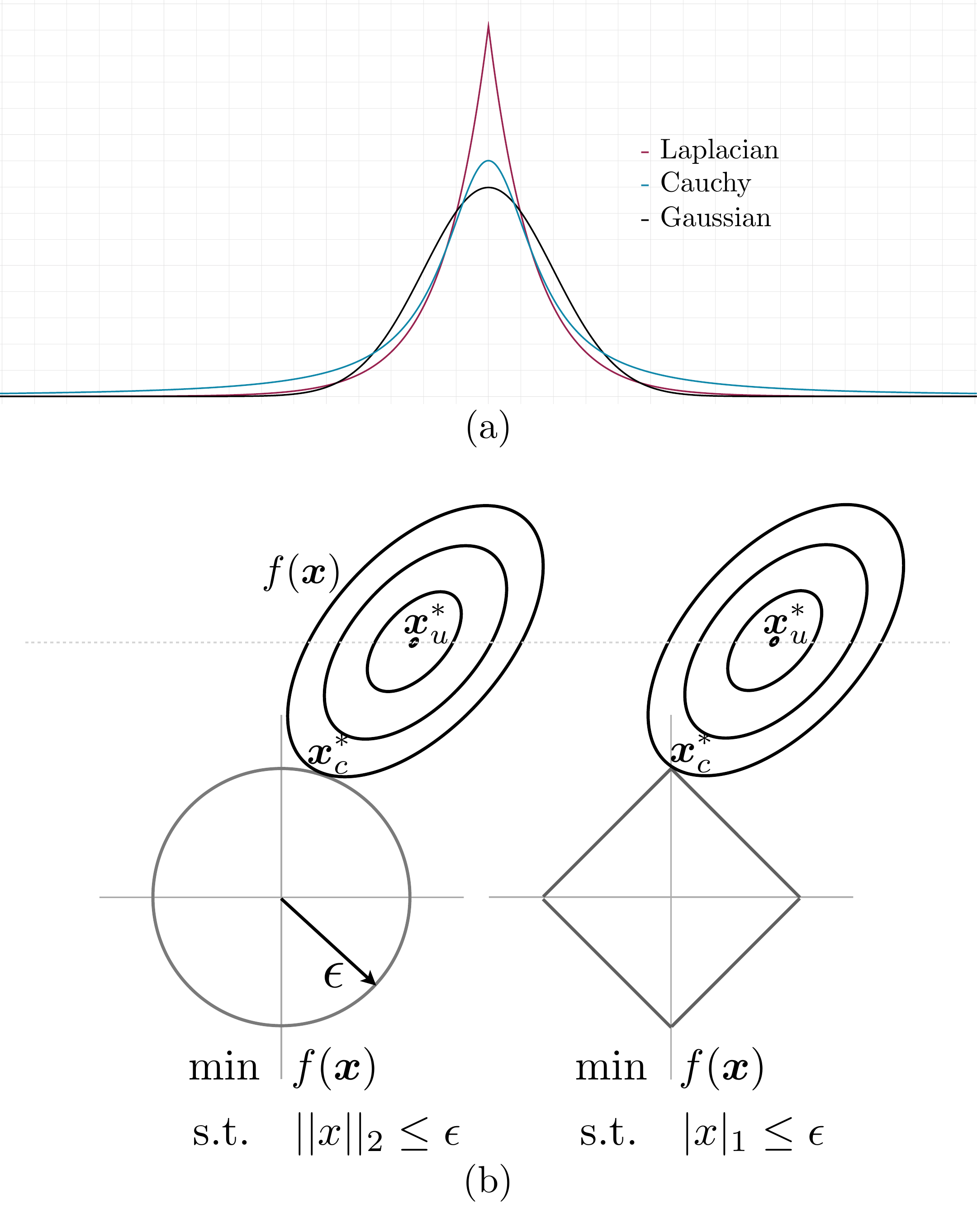}
\caption{Sparsifying distributions and their effect as regularizers. (a) For a given variance, the Laplacian and Cauchy distributions have more probability mass centered around zero. (b) As a result, $L1$ constraints favour axis-aligned solutions (only the second dimension is non-zero).
\label{fig:l1}}
\end{figure}

Tibshirani~\citep{tibshirani_JRSS_1996} introduced a convex form of the sparse inference problem - estimating Laplacian distributed coefficients which minimize a least-squares reconstruction error - using $L_1$-norm regularization and presented a method for solving it with existing quadratic programming techniques.



The full convolutional sparse coding algorithm culminated in the work of \citep{grosse_UAI_2007}. Fundamentally, Grosse extended Olshausen and Field's sparse coding algorithm to include convolutional constraints and generalized Lewicki and Sejnowski's convolutional sparse inference to 2D. Algorithmically, Grosse drew on the work of Tibshirani~\citep{tibshirani_JRSS_1996} in expressing Laplacian distributed coefficients as $L_1$-norm regularization, and used the feature sign search minimization algorithm proposed by his colleague in the same year~\citep{lee_NIPS_2007} to solve it efficiently. The form he introduced is the now canonical bilinear convolutional sparse coding algorithm.

Convolutional sparse coding has found application in learning Gabor-like bases that reflect the receptive fields of the primary visual cortex~\citep{olshausen_SCIENCE_1997}, elemental motifs of visual~\citep{zeiler_CVPR_2010}, speech~\citep{grosse_UAI_2007,lewicki_NIPS_1999} and musical~\citep{grosse_UAI_2007,morup_JOML_2008} perception, a basis for human motion and articulation~\citep{zhu_PAMI_2014}, mid-level discriminative patches~\citep{sohn_ICCV_2011} and unsupervised learning of hierarchical generative models~\citep{lee_ICML_2009,zeiler_CVPR_2010}.

The large-scale nature of the latter applications have placed great demands on the computational efficiency of the underlying algorithms. Coupled with the steady advances in machine learning and computing, this has given rise to a range of optimization approaches for convolutional sparse coding. \citep{chalasani_IJCNN_2012} introduced a convolutional extension to the FISTA algorithm for sparse inference~\citep{beck_JIS_2009}, and \citep{bristow_CVPR_2013} introduced a Fourier method based on the closely-related ADMM~\citep{boyd_book_2010}. Convolutional basis learning has largely been relegated to gradient descent, however~\citep{grosse_UAI_2007} introduced a Fourier domain adaptation that minimized the inequality constrained problem via the Lagrange dual.

In this paper, we make the following contributions:
\begin{enumerate}
\item We provide a brief discussion detailing the works that influenced the development of the convolutional sparse coding algorithm and how it stems from natural processes.
\item We argue that \emph{convolutional} sparse coding makes a set of assumptions that are more appropriate in many tasks where block or sampled sparse coding is currently used.
\item We discuss practical optimization of convolutional sparse coding, including speed, memory usage and assumptions on boundary conditions.
\item We show through a number of examples the applicability of convolutional sparse coding to a wide range of problems that arise in computer vision, and when particular problems benefit from different optimization methods.
\end{enumerate}
It is worth emphasising that the algorithms presented in this work are not new. Detailed treatments of the algorithms can be found in their respective works~\citep{bristow_CVPR_2013,chalasani_IJCNN_2012,grosse_UAI_2007}.
Rather, this paper aims to present the available algorithms in a unified framework and act as a resource for practitioners wishing to understand their properties.

\subsection{Outline}
We begin in \sec{csc} with an introduction of the convolutional sparse coding formulation. We discuss optimization methods for convolutional sparse inference first in \sec{coefficients}, as they are of most algorithmic interest, followed by methods for basis/filter updates in \sec{filters}. In \sec{app} we present a number of applications of convolutional sparse coding, each with a discussion of the preferred optimization method and implication of boundary effects. Proofs and implementation details are deferred until the appendices.

\subsection{Notation}
Matrices are written in bold uppercase (\eg $\A$), vectors in bold lowercase (\eg $\x$) and scalars in regular typeface (\eg $K$). Convolution is represented by the $\conv$ operator, and correlation by the $\corr$ operator. We treat all signals as vectors - higher-dimensional spatial or spatio-temporal signals are implicitly vectorized - however operations on these signals are performed in their original domain. We adopt this notation because (i) the domain of the signal is largely irrelevant in our analysis and the algorithms we present, and (ii) for any linear transform in the original domain, there exists an equivalent transform in the vectorized domain.

A $\hat{}$ applied applied to any vector denotes the Discrete Fourier Transform (DFT) of a vectorized signal $\a$ such that $\hat{\a} \gets \dft(\a) = \F\a$, where $\dft()$ is the Fourier transform operator and $\F$ is the $D \times D$ matrix of complex Fourier bases.

\section{Convolutional Sparse Coding}
\label{sec:csc}
The convolutional sparse coding problem consists of minimizing a convolutional model-fitting term $f$ and a sparse regularizer $g$,
\begin{alignat}{4}
\label{eqn:bilinear}
&\arg\min_{\d,\z} \quad && f(\d, \z) + \beta g(\z)
\end{alignat}
where $\d$ is the convolutional kernel, $\z$ are the set of sparse coefficients, and $\beta$ controls the tradeoff between reconstruction error and sparsity of representation. The input can be reconstructed via the convolution,
\begin{align}
\x = \d \conv \z \;.
\end{align}
Assuming Gaussian distributed noise and Laplacian distributed coefficients, \eqn{bilinear} can be written more formally as,
\begin{alignat}{4}
\label{eqn:csc}
&\arg\min_{\d,\z} \quad && \half ||\x - \d \conv \z ||^2_2 + \beta || \z ||_1 \nonumber \\
&\; \st \quad && || \d ||^2_2 \le 1
\end{alignat}
The remainder of our analysis is based around efficient methods of optimizing this objective. The objective naturally extends to multiple images and filters,
\begin{alignat}{4}
&\arg\min_{\d,\z} \quad && \half \sum_{i=1}^M || \x_i - \sum_{j=1}^N (\d_j \conv \z_{i,j}) ||^2_2 + \beta \sum_{i=1}^M \sum_{j=1}^N || \z_{i,j} ||_1 \nonumber \\
&\; \st \quad &&  || \d_j ||^2_2 \le 1 \;\; \forall \; j \in 1 \dots N 
\end{alignat}
however this form quickly becomes unwieldy, so we only use it when the number of filters or images needs to be emphasized.

Contrast the objective of \eqn{csc} with that of conventional sparse coding,
\begin{alignat}{4}
\label{eqn:sc}
&\arg\min_{\B,\Z} \quad && ||\X - \B \Z ||^2_2 + \beta || \Z ||_1 \nonumber \\
&\; \st \quad && || \B_i ||^2_2 \le 1 \;\; \forall \; i
\end{alignat}
Here, we are solving for a set of basis vectors $\B$ and sparse coefficients $\Z$ in alternation that reconstruct patches or samples of the signal $\X$ \emph{in isolation}. This is an important distinction to make, and forms the fundamental difference between convolutional and patch-based sparse coding. 

In the limit when $\X$ contains \emph{every patch} from the full image, the two sparse coding algorithms behave equivalently, however the patch-based algorithm must store a redundant amount of data, and cannot take advantage of fast methods for evaluating the inner product of the basis with each patch (\ie convolution).

The objective of \eqn{csc} is bilinear -- solving for each variable whilst holding the other fixed yields a convex subproblem, however the objective is not jointly convex in both. We optimize the objective in alternation, iterating until convergence. There are no guarantees that the final minima reached is the global minima, however in practice multiple trials reach minima of comparable quality, even if the exact bases and distribution of coefficients learned are slightly different.

In the sections that follow, we introduce a range of algorithms that can solve for the filters and coefficients. Since the alternation strategy treats each independently, we largely consider the algorithms in isolation, however some care must be taken in matching appropriate boundary condition assumptions.

\subsection{Other Formulations}
It should be noted that the algorithm of \eqn{csc} is not the only conceivable formulation of convolutional sparse coding. In particular, there has been growing interest in non-convex sparse coding with hyper-Laplacian, $L_{0+}$ and other exotic priors~\citep{zhang_AS_2010,wipf_IT_2011}. Whilst these methods have not yet been extended to \emph{convolutional} sparse coding, there is no fundamental barrier to doing so.

\section{Solving for Coefficients}
\label{sec:coefficients}
It is natural to begin with methods for convolutional sparse inference, since this is where most of the research effort has been focussed. Solving for the coefficients $\z$ involves optimizing the unconstrained objective,
\begin{alignat}{4}
& \arg\min_{\z} || \x - \d \conv \z ||^2_2 + \beta || \z ||_1
\end{alignat}
where $\beta$ controls the tradeoff between sparsity and reconstruction error. This objective is difficult to solve because (i) it involves a non-smooth regularizer, (ii) the least-squares system cannot be solved directly (forming explicit convolutional matrices for large inputs is infeasible), and (iii) the system involves a large number of variables, especially if working with megapixel imagery, \etc.

In the case of multiple images and filters,
\begin{alignat}{4}
\label{eqn:coefficients-multiple}
&\arg\min_{\z} \;\;&&\half \sum_{i=1}^M || \big( \x_i - \sum_{j=1}^N (\d_j \conv \z_{i,j}) \big) ||^2_2 \nonumber \\ 
&&&+ \beta \sum_{i=1}^M \sum_{j=1}^N || \z_{i,j} ||_1
\end{alignat}
and given that the minima of the sum of convex functions is the sum of their minima,
\begin{alignat}{4}
&\min_{\z} \sum_{i=1}^M f_i (\z) \;=\; \sum_{i=1}^M \min_{\z} f_i (\z)
\end{alignat}
the coefficients for each image can be inferred separately,
\begin{align}
&\z^\star_i = \arg\min_{\z} \;\; \half || \x_i - \sum_{j=1}^N ( \d_j \conv \z_{i,j} ) ||^2_2 + \beta \sum_{j=1}^N || \z_{i,j} ||_1
\end{align}
The two methods that we introduce are both based around partitioning the objective into the smooth model-fitting term and non-smooth regularizer that can then be handled separately.

\subsection{ADMM Partitioning}
\label{sec:coefficients-admm}
The alternating direction method of minimizers (ADMM), was proposed jointly by~\citep{glowinski_RFA_1975,gabay_CMA_1976}, though the idea can be traced back as early as Douglas-Rachford splitting in the mid-1950s. \citep{boyd_book_2010} presents a thorough overview of ADMMs and their properties. ADMMs solve problems of the form,
\begin{alignat}{4}
& \arg\min_{\z, \t} \quad && g(\z) + h(\t) \nonumber \\
& \;\st                                     && \A\z + \B\t =  \c
\end{alignat}

To express convolutional sparse inference in this form, we perform the (somewhat unintuitive) substitution,
 \begin{alignat}{4}
\label{eqn:admm-g}
& g(\z) = \half || \x - \d \conv \z ||^2_2 \;\;\;\;\;\;\;\;&& h(\z) = \beta || \z ||_1 \\
& \A = \I \;\;\;\;\;\;\;\;\;\;\; \B = -\I && \c = 0
\end{alignat}
to obtain,
 \begin{alignat}{4}
 &\arg\min_{\z,\t} \quad && \half || \x - \d \conv \z ||^2_2 + \beta || \t ||_1 \nonumber\\
 &\;\st && \z = \t
 \end{alignat}
By introducing a proxy $\t$, the loss function can be treated as a sum of functions of two independent variables, and with the addition of equality constraints, the minima of the new constrained objective is the same as the original unconstrained objective.
 
Whilst the individual functions may be easier to optimize, there is added complexity in enforcing the equality constraints. Taking the Lagrangian of the augmented objective,
\begin{alignat}{2}
& \L(\z, \t, \l) = g(\z) + h(\t) + \frac{\rho}{2} || \z - \t + \l ||^2_2
\end{alignat}
the solution is to minimize the primal variables, and maximize the dual variables,
\begin{alignat}{2}
&\z^\star, \t^\star, \l^\star \;=\; \arg\max_{\l} \left( \arg\min_{\z,\t} \L(\z,\t,\l) \right)
\end{alignat}
Optimizing in alternation (which accounts for the term \emph{alternating direction}) yields a strategy for updating the variables involving a function of a single variable plus a proximal term binding all variables,
\begin{alignat}{4}
\label{eqn:admm}
&\z^{k+1} &&= \arg\min_{\z} \;\; g(\z^k) + \frac{\rho}{2} || \z^k - (\t - \l) ||^2_2 \\
&\t^{k+1} &&= \arg\min_{\t} \;\; h(\t^k) + \frac{\rho}{2} || \t^k - (\z + \l) ||^2_2 \\
&\l^{k+1} &&= \l^k + \rho (\z - \t)
\end{alignat}

The intuition behind this strategy is to find a set of model parameters which don't deviate too far from the regularization parameters, then visa versa, to find a set of regularization parameters which are close to the model parameters. The Lagrange variables impose a linear descent direction that force the two primal variables to equality over time.

For this strategy to be effective, the sum of the function $g$ or $h$ and an isotropic least-squares must be easy to solve.

Substituting the convolutional sparse inference objective into \eqn{admm},
\begin{alignat}{4}
\label{eqn:admm-subproblems}
&\z^{k+1} &&= \arg\min_{\z} \half || \x - \d \conv \z^k ||^2_2 + \frac{\rho}{2} || \z^k - (\t - \l) ||^2_2 \hspace{-10mm} \\
&\t^{k+1} &&= \arg\min_{\t} \beta || \t^k ||_1 + \frac{\rho}{2} || \t^k - (\z + \l) ||^2_2
\end{alignat}
The $\z$ update takes the form of generalized Tikhonov regularization. In the $\t$ update, the $L_1$-regularizer can now be treated independently of the model-fitting term, and importantly the added proximal term is an isotropic Gaussian, so each pixel of $\t$ can be updated independently,
\begin{alignat}{4}
&t^{k+1} &&= \arg\min_{t} \;\; \beta | t^k | + \frac{\rho}{2} (t^k - z + \lambda)^2
\end{alignat}
the solution to which is the soft thresholding operator,
\begin{alignat}{4}
\label{eqn:soft}
&t^\star = \sgn \left(z + \lambda \right) \cdot \max \left\{ | z + \lambda | - \frac{\beta}{\rho}, 0 \right\}
\end{alignat}

In the ADMM, the minimizer $\t^\star$ is exactly sparse, whilst the minimizer $\z^\star$ is only close to sparse. If exact sparsity is a concern in the problem domain, $\t^\star$ should be retained at the point of convergence.

\subsection{FISTA/Proximal Gradient}
Proximal gradient methods are a close parallel to \mbox{ADMMs}. They generalize the problem of projecting a point onto a convex set, and often admit closed-form solutions. \citep{boyd_2013} present a thorough review of proximal algorithms and their relation to \mbox{ADMMs}. The convolutional Lasso problem introduces the splitting,
\begin{alignat}{4}
\label{eqn:proximal}
& g(\z) = \half || \x - \d \conv \z ||^2_2 \;\;\;\;\;\;\;\; h(\z) = \beta || \z ||_1
\end{alignat}
with gradient and proximal operator,
\begin{alignat}{4}
\label{eqn:proximal-gradient}
& \grad g(\z) = \D^T(\D\z - \x) \;\;\;\;\;\;\;\; \prox_h(\z) = \S(\z)
\end{alignat}
where $\S$ is the soft thresholding operator of \eqn{soft}. In this case, the proximal algorithm is finding the closest minimizer to the convolutional least squares model fitting term that projects onto the $L_1$-ball (with radius proportional to $\beta$).

The FISTA algorithm of~\citep{beck_JIS_2009} presents an efficient update strategy for solving this problem by incorporating an optimal first-order method in the gradient computation (discussed in \sec{agd}).
 
FISTA and ADMMs both have similar computational complexity, each requiring updates to a least-squares problem and evaluation of a soft thresholding operator. One disadvantage of FISTA is the requirement of gradient-based updates to the functional term. ADMMs, on the other hand, make a more general set of assumptions, requiring only that the objective value in each iteration is reduced. In cases where solving the objective is similar in complexity to evaluating the gradient, ADMMs may converge faster.

\subsection{Iterative Optimziation}
So far we have neglected to show how convolution in \eqn{admm-g} and \eqn{proximal} is actually performed.

The classical approach to convolution is to assume Dirichlet boundary conditions, \ie that values outside the domain of consideration are zero. In such a case $\x \in \R^{P \times Q}$, $\d \in \R^{R \times S}$ and $\z \in \R^{P \times Q}$. Another approach is to take only the convolution between the fully overlapping portions of the signals -- `valid' convolution -- which results in $\z \in \R^{(P+R-1) \times (Q+S-1)}$.

In both approaches, the signals \emph{must} be convolved in the spatial domain, which is an $\O(PQRS)$ operation.

One way to alleviate the computational cost is to assume periodic extension of the signal, where convolution is then diagonalized by the Fourier transform,
\begin{align}
& \d \conv \z = \dft^{-1} \left\{ \dft(\d) \cdot \dft(\z) \right\}
\end{align}
where $\d \in \R^{P \times Q}$ and $\z \in \R^{P \times Q}$ (see \sec{padding} for the correct method of padding $\d$ to size). This method of convolution has cost $\O(PQ\log(PQ))$.

Both ADMMs and FISTA can take advantage of iterative methods, by taking accelerated (proximal) gradient steps. In the next section we show how solving the entire system of equations in the Fourier domain is only slightly more complex than performing a single gradient step, and can lead to faster overall convergence of the ADMM. An important distinction between FISTA and ADMMs is that FISTA is constrained to use gradient updates -- it cannot take advantage of direct solvers.

\subsection{Direct Optimization}
\label{sec:coefficients-direct}
Given Dirichlet boundary assumptions, a filter kernel $\d$ can be represented (in 2D) as a block-Toeplitz-Toeplitz matrix such that $g(\z)$ of \eqn{proximal} becomes,
\begin{align}
&g(\z) = \half || \x - \D \z ||^2_2
\end{align}
where,
\begin{align}
& \D = \left[ \D_1 \dots \D_N \right]
\;\;\;\;\;\;\;
\z = \left[ \begin{array}{c}
\z_1 \\ \vdots \\ \z_N
\end{array} \right]
\end{align}
which is the canonical least-squares problem. There are a plethora of direct solvers for this problem, however unlike Toeplitz matrices, there are no known fast \mbox{($< \O(n^2)$)} methods for inverting block-Toeplitz-Toeplitz matrices, so general purpose solvers must be used. This requires constructing the full $\D$. For a $P \times Q$ input, and $N$ filters each of support $R \times S$, the matrix $\D$ will have $\O(NPQRS)$ non-zero values.

One way to alleviate the computational cost is to assume periodic extension of the signal, where convolution is then diagonalized by the Fourier transform.

This provides an efficient strategy for inverting the system,
\begin{align}
\label{eqn:fourier-least-squares}
& g(\zf) = \half || \xf - \Df \zf ||^2_2
\end{align}
where,
\begin{align}
& \Df = \left[ \diag(\df_1) \dots \diag(\df_N) \right]
\;\;\;\;\;
\zf = \left[ \begin{array}{c}
\zf_1 \\ \vdots \\ \zf_N
\end{array} \right]
\end{align}

The equivalence between \eqn{proximal} and \eqn{fourier-least-squares} relies upon Parseval's theorem,
\begin{align}
& \x^T \x = K \xf^T \xf
\end{align}
where $K$ is a constant scaling factor between the domains. Since the $L_2$ norm is rotation invariant, the minimizer of \eqn{fourier-least-squares} is the Fourier transform of the minimize of \eqn{proximal}.

The system of \eqn{admm-subproblems} can be solved directly in an efficient manner in the Fourier domain by observing that each frequency band in a single image $\x$ is related only by the same frequency band in each of the filters and coefficients, \eg,
\begin{align}
\xf_{1,1} = \df_{1,n} \zf_{n,1} \;\;\; \forall \; n \in 1 \dots N
\end{align}

Finding the optima of $\zf$ across all $n$ channels for frequency band $1$ now involves
\begin{align}
\label{eqn:coefficients-dense}
&\zf_{n,1} = \left( \df_{1,n}^T \df_{1,n} + \rho \I_{n,1} \right)^{-1} \left( \xf_{1,1} \df_{1,n}^T + \rho (\tf_{n,1} - \lf_{n,1}) \right)
\end{align}

Note that $\df_{1,n}^T \df_{1,n}$ is a rank-1 matrix. Thus from the matrix inversion lemma  (\app{inversion_lemma}), 
\begin{align}
&\left( \df_{1,n}^T \df_{1,n} + \rho \I_{n,1} \right) \hspace{-4mm}&&= \frac{1}{\rho} \left( \I_{n,1} - \frac{1}{\rho + \df_{1,n} \df_{1,n}^T} \right) \df_{1,n}^T \df_{1,n} \nonumber \\
&&&= \frac{1}{\rho} ( \I_{n,1} - K) \df_{1,n}^T \df_{1,n}
\end{align}
where $K$ is a scalar, since $\d_{1,n} \d_{1,n}^T$ is a scalar. The block of $\z_{n,1}$ can thus be solved by,
\begin{align}
& \zf_{n,1} = \frac{1}{\rho} ( \I_{n,1} - K) \df_{1,n}^T \df_{1,n} \left( \xf_{1,1} \df_{1,n}^T + \rho (\tf_{n,1} - \lf_{n,1}) \right)
\end{align}
which is just a series of multiplications, so a solution can be found in $\O(n^2)$ time -- no inversion is required.

The assumption of periodic extension is sometimes not indicative of the structure observed in the signal of interest, and can cause artefacts along the boundaries. In such a case, a more sensible assumption is to assume Neumann boundary conditions, or \emph{symmetric} reflection across the boundary. This assumption tends to minimize boundary distortion for small displacements. The resulting matrix can be diagonalised by the discrete cosine transform (DCT). There exists a generalization of Parseval's theorem that extends to the DCT, however some care must be taken with understanding the nuances between the 16 types of DCT. \citep{martucci_1993} provides a comprehensive exposé on the topic.

\section{Solving for Filters}
\label{sec:filters}
Solving for the filters $\d$ involves optimizing,
\begin{alignat}{4}
\label{eqn:basis}
&\arg\min_{\d} \quad && \half ||\x - \d \conv \z ||^2_2 \nonumber \\
&\; \st \quad && || \d ||^2_2 \le 1
\end{alignat}
This is a classical least squares objective with norm inequality constraints. Norm constraints on the bases are necessary since there always exists a linear transformation of $\d$ and $\z$ which keeps $(\d \conv \z)$ unchanged whilst making $\z$ approach zero. Inequality constraints are sufficient since deflation of $\z$ will always cause $\d$ to lie on the constraint boundary $||\d||^2_2 = 1$ (whilst forming a convex set).

\eqn{basis} is a quadratically constrained quadratic program (QCQP), which is difficult to optimize in general. Each of the following methods relax this form in one way or another to make it more tractable to solve.

The convolutional form of the least squares term also poses some challenges for optimization, since (i) the filter has smaller support than the image it is being convolved with, and (ii) forming an explicit multiplication between the filters and a convolutional matrix form of the images is prohibitively expensive (as per \sec{coefficients-direct}).

In the case of multiple images and filters,
\begin{alignat}{4}
&\arg\min_{\d} \quad && \half || \sum_{i=1}^M \big( \x_i - \sum_{j=1}^N (\d_j \conv \z_{i,j}) \big) ||^2_2 \nonumber \\
&\; \st \quad &&  || \d_j ||^2_2 \le 1 \;\; \forall \; j \in 1 \dots N 
\end{alignat}
one can see that all filters are jointly involved in reconstructing the inputs, and so must be updated jointly.

\subsection{Gradient Descent}
The gradient of the objective is given by,
\begin{align}
\label{eqn:basis_grad}
\grad f_i = \left( \sum_{j=1}^N \d_j \conv \sum_{k=1}^M ( \z_{i,j} \corr \z_{i,k} ) \right) - \sum_{k=1}^M \left(  \x_{i} \corr \z_{i,j} \right)
\end{align}
This involves collecting the correlation statistics across the coefficient maps and images. Since the filters are of smaller support than the coefficient maps and images, we collect only ``valid'' statistics, or regions that don't incur boundary effects. In the autocorrelation of $\z$, one of the arguments must be zero padded to the appropriate size.

Given the gradient direction, updating the the filters involves,
\begin{alignat}{4}
& \d^{k+1} \;=\; \d^{k} - t\grad f
\end{alignat}
Computing the step size, $t$, can be done either via line search or by solving a 1D optimization problem which minimizes the reconstruction error in the gradient direction,
\begin{alignat}{4}
&\arg\min_{t} \;\;&& || \x - \underbrace{\left( \d^{k} - t\grad f \right)}_{\d^{k+1}} \conv z ||^2_2
\end{alignat}
the closed-form solution to which is,
\begin{alignat}{4}
& t \; = \; \frac{(\x - \d \conv \z)^T (\grad f \conv \z)}{(\grad f \conv \z)^T (\grad f \conv \z)}
\end{alignat}
and involves evaluating only $2N$ convolutions ($N$ if the $\x - \d \conv \z$ term has been previously computed as part of a stopping criteria, \etc). In the case of a large number of inputs $M$, stochastic gradient descent is typically used~\citep{mairal_ICML_2009}.

Typically after each gradient step, if the new iterate exists outside the $L_2$ unit ball, the result is projected back onto the ball. This is not strictly the correct way to enforce the norm-constraints, however it works without side-effects in practice. The ADMM (\sec{basis-admm}) and Lagrange dual (\sec{ld}) methods on the other hand, both solve for the norm constraints exactly.   

\subsection{Nesterov's Accelerated Gradient Descent}
\label{sec:agd}
Prolific mathematician Yurii Nesterov introduced an optimal\footnote{Optimal in the sense that it has a worst-case convergence rate that cannot be improved whilst remaining first-order.} first-order method for solving smooth convex functions~\citep{Nesterov1983}. Convolutional least squares objectives require a straightforward application of accelerated proximal gradient (APG).

Without laboring on the details introduced by Nesterov, the method involves iteratively placing an isotropic quadratic tangent to the current gradient direction, then shifting the iterate to the minima of the quadratic. The curvature of the quadratic is computed by estimating the Lipschitz smoothness of the objective.

Checking for Lipschitz smoothness feasibility using backtracking requires two projections per iteration. This can be costly since each projection involves multiple convolutions. In practice we find this method no faster than regular gradient descent with optimal step-size calculation.

\subsection{ADMM Partitioning}
\label{sec:basis-admm}
As per \sec{coefficients-admm}, treating convolution in the Fourier domain can lead to efficient direct optimization. Unlike solving for the coefficients, however, the learned filters must be constrained to be small support, and there is no way to do this explicitly via Fourier convolution.

An approach to handling this is via ADMMs again,
\begin{alignat}{4}
& \arg\min_{\d,\s} \;\;&& \half || \xf - \Zf \sf ||^2_2 \nonumber \\
& \;\st && \d_j = \Phib^T \sf_j \;\;\;\; \forall \; j \in 1 \dots N \nonumber \\
          &&& || \d_j ||^2_2 \le 1 \;\;\;\;\; \forall \; j \in 1 \dots N
\end{alignat}
where,
\begin{align}
&\xf = \left[\begin{array}{c}
\xf_1 \\ \vdots \\ \xf_M
\end{array} \right]
\;\;\;\;\;\;\;
\df = \left[\begin{array}{c}
\df_1 \\ \vdots \\ \df_N
\end{array} \right]
\nonumber \\
&\Zf = \left[\begin{array}{ccc}
\diag(\zf_{1,1}) & \dots & \diag(\zf_{1,N}) \\
\vdots & \ddots & \\
\diag(\zf_{M,1}) && \diag(\zf_{M,N})
\end{array} \right]
\end{align}
and $\Phib$ is a submatrix of the Fourier matrix that corresponds to a small spatial support transform.

Intuitively, we are trying to learn a set of filters $\sf$ that minimize reconstruction error in the Fourier domain and are small support in the spatial domain.

Taking the augmented Lagrangian of the objective and optimizing over $\d$ and $\s$ in alternation yields the update strategy,
\begin{alignat}{6}
& \sf^{k+1} &=&\arg\min_{\sf} \; && || \xf - \Zf \sf^k ||^2_2 + \frac{\rho}{2} || \sf^k - (\Phib \d - \lf) ||^2_2 \hspace{-5mm}\\
& \d^{k+1} &=& \arg\min_{\d} && || \Phib \d^k - (\sf + \lf) ||^2_2 \nonumber \\
&&& \; \st && || \d_j ||^2_2 \le 1 \;\;\;\;\; \forall \; j \in 1 \dots N \\
&\l^{k+1} &=& \;\l^k + \rho (\Phib \d - \sf) \span\omit \hspace{-20mm}
\end{alignat}

In a similar fashion to \eqn{coefficients-dense}, each frequency component in $\sf$ jointly across all filters can be solved for independently via a variable reordering to produce $P \times Q$ dense systems of equations.

Solving for $\d$ appears more involved, however the unconstrained loss function can be minimized in closed-form,
\begin{alignat}{4}
&\d^\star &&= \d^T \Phib^T \Phib \d - 2 \d^T \Phib^T (\sf - \lf) + c \nonumber \\
&&&= \Phib^T (\sf - \lf)
\end{alignat}
since $\Phib$ is an orthonormal matrix, and thus $\Phib^T \Phib = \I$. Further, the matrix multiplication can instead be replaced by the inverse Fourier transform, followed by a selection operator $\mathcal{M}$ which keeps only the small support region,
\begin{align}
& \d^\star = \mathcal{M} \left( \dft^{-1} \{ \sf - \lf \} \right)
\end{align}

Handling the inequality constraints is now trivial. Since $\Phib$ is orthonormal (implying an isotropic regression problem), projecting the optimal solution to the unconstrained problem onto the $L_2$ ball,
\begin{align}
& \d^\star = \left\{ \begin{array}{ll}
\;\;|| \d^\star_k ||_2^{-2} \d^\star_k, & \;\;\;\;\mbox{if } || \d^\star_k ||^2_2 \ge 1 \\
\;\;\d^\star_k, & \;\;\;\;\mbox{otherwise}
\end{array} \right.
\end{align}
 is \emph{equivalent} to solving the constrained problem.

\subsection{Lagrange Dual}
\label{sec:ld}
Rather than using gradient descent with iterative projection, \eqn{basis} can be solved by taking the Lagrange dual in the Fourier domain,
\begin{align}
& \L(\df, \lf) = || \xf - \Zf \d ||^2_2 + \df^T \boldsymbol{\Lambda} \df - 1^T \l
\end{align}
where $\l \ge 0$ are the dual variables, and $\boldsymbol{\Lambda} = \diag(\l)$.
Finding the optimum involves minimizing the primal variables, and maximizing the dual variables,
\begin{align}
\df^\star, \lf^\star = \arg\max_{\lf} \left( \arg\min_{\df} \L(\df, \lf) \right)
\end{align}
The closed-form solution to the primal variables is,
\begin{align}
&\df^\star = (\Zf^T \Zf + \boldsymbol{\Lambda})^{-1} (\Zf^T \xf)
\end{align}
Substituting this expression for $\df$ into the Lagrangian, we analytically derive the dual optimization problem.

\subsection{Convolving with Small Support Filters in the Fourier Domain}
\label{sec:padding}
\begin{figure}
\includegraphics[width=\columnwidth]{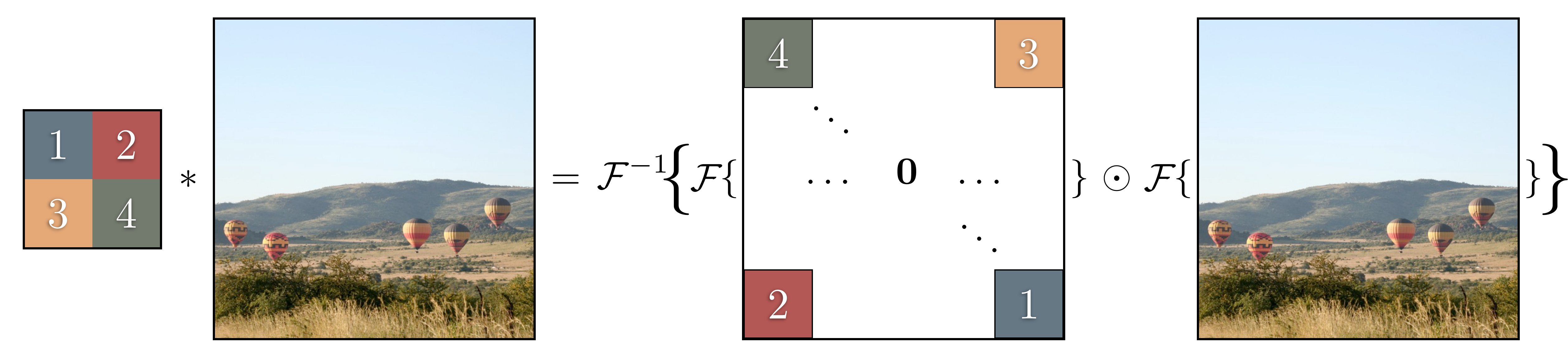}
\caption{Swapping quadrants and padding the filter to the size of the image permits convolution in the Fourier Domain
\label{fig:fourier-padding}}
\end{figure}

In order to convolve two signals in the Fourier domain, their lengths must commute. This involves padding the shorter signal to the length of the longer. Some care must be taken to avoid introducing phase shifts into the response, however. Given a 2D filter $\z \in \R^{P,Q}$, we can partition it into $4$ blocks,
\begin{align}
\z = \left | \begin{array}{cc}
\z_{1,1} & \z_{1,2} \\
\z_{2,1} & \z_{2,2}
\end{array}
\right |
\end{align}
where, in the case of odd-sized filters, the blocks are partitioned \emph{above and to the left} of the central point. Given a 2D image $\x \in \R^{M,N}$, the padded representation $\z^\star \in \R^{M,N}$ can thus be formed as,
 \begin{align}
\z = \left | \begin{array}{ccccc}
\z_{2,2} &&&& \z_{2,1} \\
& \ddots &&& \\
& \dots & \boldsymbol{0} & \dots & \\
&&& \ddots & \\
\z_{1,2} &&&& \z_{1,1}
\end{array}
\right |
\end{align}
This transform is illustrated in \fig{fourier-padding}. For a comprehensive guide to Fourier domain transforms and identities, including appropriate handling of boundary effects and padding, see~\citep{oppenheim_book_1996}. 

\section{Stopping Criteria}
For the gradient-based algorithms -- gradient descent, APG and FISTA -- a sufficient stopping criteria is to threshold the residual between two iterates,
\begin{alignat}{2}
& || \x^{k+1} - \x^{k} ||^2_2 \le \epsilon
\end{alignat}
where $\x$ is the variable being minimized.

Estimating convergence of the ADMM-based methods is more involved, including deviation from primal feasibility,
\begin{alignat}{4}
& & || \d - \s ||^2_2 \le \epsilon \;,\;\;\;\;\;\;\;\; && || \z - \t ||^2_2 \le \epsilon
\end{alignat}
and dual feasibility,
\begin{alignat}{2}
& || \l^{k+1} - \l^{k} ||^2_2 \le \epsilon
\end{alignat}
However, in practice it is usually sufficient to measure only primal feasibility, since if the iterates have reached primal feasibility, dual feasibility is unlikely to improve (this is doubly true when using a strategy for increasing $\rho$).

\section{Applications}
\label{sec:app}
\subsection{Example 1 - Image and Video Compression}
Many image and video coding algorithms such as JPEG~\citep{wallace_1991} and H.264~\citep{sullivan_2004} discretize each image into blocks which are transformed, quantized and coded independently. 

Using convolutional sparse coding, an entire image $\x$ can be coded onto a basis $\d$ with sparse reconstruction coefficients $\z$. Quality and size can be controlled via the $\beta$ parameter. The basis can either be specific to the image, or a generic basis (such as Gabor) which is part of the coding spec and need not be transferred with the image data.

Since the coefficients of $\z$ are exactly sparse by virtue of the soft-thresholding operator, the representation can make effective use of run-length and Huffman entropy coding techniques. \footnote{JPEG also uses run-length coding but its efficiency is a function of the quantization artefacts.}

To reconstruct the image $\x_r$, the decoder simply convolves the bases with the transmitted coefficients,
\begin{alignat}{2}
& \x_r = \sum_{j=1}^N \d_j \conv \z_j
\end{alignat}
This matches the media model well: media is consumed more frequently than it is created, so encoding can be costly (in this case an inverse inference problem) but decoding should be fast -- convolutional primitives are hardware-accelerated on almost all modern chipsets.

\subsection{Example 2 - A basis for Natural Images}
\label{sec:gabor}
The receptive fields of simple cells in the mammalian primary visual cortex can be characterized as being spatially localized, oriented and bandpass. \citep{olshausen_field_NATURE_1996} hypothesized that such fields could arise spontaneously in an unsupervised strategy for maximizing the sparsity of the representation. Sparsity can be interpreted biologically as a metabolic constraint - firing only a few neurons in response to a stimulus is clearly more energy efficient than firing a large number.

\citep{olshausen_field_NATURE_1996} use traditional patch-based sparse coding to solve for a set of basis functions. The famous result is that the basis functions resemble Gabor filters. However, a large number of the bases learned are translations of others -- an artefact of sampling and treating each image patch independently.

It is well-understood that the statistics of natural images are translation invariant~\citep{hyvarinen_2009}, \ie the covariance of natural images depends only on the distance,
\begin{align}
\Sigma \left( \I(x,y), \; \I(x', y') \right) = f\left( (x - x')^2 + (y - y')^2 \right)
\end{align}
Thus it makes sense to code natural images in a manner that does not depend on exact position of the stimulus. Convolutional sparse coding permits this, and as a result produces a more varied range of basis elements than simple Gabor filters when coding natural images. The sparsity pattern of convolutional coefficients also has a mapping onto the receptive fields of active neurons in V1.\footnote{Unlike many higher regions within the visual cortex, V1 is retinotopic -- the spatial location of stimulus in the visual world is highly correlated with the spatial location of active neurons within V1.}

\subsection{Example 3 - Structure from Motion}
Trajectory basis Non-Rigid Structure from Motion (NRSfM) refers to the process of reconstructing the motion of 3D points of a non-rigid object from only their 2D projected trajectories.

Reconstruction relies on two inherently conflicting factors: (i) the condition of the composed camera and trajectory basis matrix, and (ii) whether the trajectory basis has enough degrees of freedom to model the 3D point trajectory. Typically, (i) is improved with a low-rank basis, and (ii) is improved with a higher-rank basis. 

The Discrete Cosine Transform (DCT) basis has traditionally been used as a generic basis for encoding motion trajectories, however choosing the correct rank has been a difficult problem. \citep{zhu_PAMI_2014} proposed the use of convolutional sparse coding to learn a compact basis that could model the trajectories taken from a corpus of training data.

Learning the basis proceeds as per usual,
\begin{alignat}{4}
&\arg\min_{\d,\z} \quad && \half \sum_{i=1}^M ||  \big( \x_i - \sum_{j=1}^N (\d_j \conv \z_{i,j}) \big) ||^2_2 \nonumber \\
&&& + \beta \sum_{i=1}^M \sum_{j=1}^N || \z_{i,j} ||_1 \nonumber\\
&\; \st \quad &&  || \d_j ||^2_2 \le 1 \;\; \forall \; j \in 1 \dots N 
\end{alignat}
where each $\x_i$ is a 1D trajectory of arbitrary length, $\d$ the trajectory basis being learned, and $\z$ the sparse reconstruction coefficients.

Given the convolutional trajectory basis $\d$, reconstructing the sparse coefficients for the $3D$ trajectory from $2D$ observations involves,
\begin{alignat}{4}
&\z^* =&& \arg\min_{\z} \;\; && || \z ||_1 \nonumber \\
        &&& \;\st && \underbrace{\Q\x}_{\u} = \Q \sum_{j=1}^N \d_j \conv \z_j
\end{alignat}
where $\u$ are the $2D$ observations of the $3D$ points $\x$ that have been imaged by the camera matrices $\Q$, one for each frame in the trajectory,
\begin{align}
\Q = \left[ 
\begin{array}{ccc}
\Q_1 && \\
& \ddots & \\
&& \Q_F
\end{array}
\right ]
\end{align}

In single view reconstruction, back-projection is typically enforced as a constraint, and the objective is to minimize the number of non-zero coefficients in the reconstructed $3D$ trajectory that satisfy this constraint.

A convolutional sparse coded basis produces less $3D$ reconstruction error than previously explored bases, including one learned from patch-based sparse coding, and a generic DCT basis. This illustrates convolutional sparse coding's ability to learn low rank structure from misaligned trajectories stemming from the same underlying dynamics (\eg articulated human motion).

\subsection{Example 4 - Mid-level Generative Parts}
Zeiler~\etal show how a cascade of convolutional sparse coders can be used to build robust, unsupervised mid-level representations, beyond the edge primitives of \sec{gabor}.

The convolutional sparse coder at each level of the hierarchy can be defined as,
\begin{alignat}{4}
& C_l(\d^l, \z^l) \;\;&&= \half \sum_{i=1}^M || \underbrace{f_s(\z^{l-1}_{i})}_{\x_i} - \sum_{j=1}^M (\d^{l}_{j} \conv \z^{l}_{i,j}) ||^2_2 \nonumber \\
&&&+ \beta \sum_{i=1}^M \sum_{j=1}^N || \z^{l}_{i,j} ||_1
\end{alignat}
where the extra superscripts on each element indicate the layer to which they are native.

The inputs $\x$ to each layer are the sparse coefficients of the previous layer, $\z^{l-1}$ after being passed through a pooling/subsampling operation $f_s(\cdot)$. For the first layer, $\z^{l-1} = \x$, \ie the input image.

The idea behind this coding strategy is that structure within the signal is progressively gathered at a higher and higher level, initially with edge primitives, then mergers between these primitives into line-segments, and eventually into recurrent object parts.

Layers of convolutional sparse coders have also been used to produce high quality latent representations for convolutional neural networks~\citep{lee_ICML_2009,kavukcuoglu_NIPS_2010,chen_PAMI_2013}, though fully-supervised back-propagation across layers has become popular more recently~\citep{krizhevsky_NIPS_2012}.

\subsection{Example 5 - Single Image Super-Resolution}
\label{sec:super-resolution}
Single-Image Super Resolution (SISR) is the process of reconstructing a high-resolution image from an observed low-resolution image. SISR can be cast as the inverse problem,
\begin{alignat}{2}
& \y = \D \B \x
\end{alignat}
where $\x$ is the latent high-resolution image that we wish to recover, $\B$ is an anti-aliasing filter, $\D$ is a downsampling matrix and $\y$ is the observed low-resolution image. The system is underdetermined, so there exist infinitely many solutions to $\x$. A strategy for performing SISR is via a straightforward convolutional extension of~\citep{yang_TIP_2010},
\begin{alignat}{4}
& \arg\min_{\mathclap{\d_L, \d_H, \z}} \quad && \sum_{i=1}^M || \D\B\x_i - \sum_{j=1}^N (\d_{L,j} \conv \D\z_{i,j}) ||^2_2 \nonumber \\
&&& + || \x_i - \sum_{j=1}^N (\d_{H,j} \conv \z_{i,j}) ||^2_2 \nonumber\\
&&& + \beta \sum_{i=1}^M \sum_{j=1}^M || \z_{i,j} ||_1 \nonumber \\
& \st && || \d_{L,j} ||^2_2 \le 1 \;\;\; \forall \; j \in 1\dots M \nonumber \\
       &&& || \d_{H,j} ||^2_2 \le 1\;\;\; \forall \; j \in 1\dots M
\end{alignat}
where $\x_i$ and $\D\B\x_i$ are a high-resolution and derived low-resolution training pair, $\D$ is the downsampling filter as before and $\z$ are a common set of coefficients that tie the  two representations together. The dictionaries $\d_L$ and $\d_H$ learn a mapping between low- and high-resolution image features. 

Given a new low-resolution input image $\x_L$, the sparse coefficients are first inferred with respect to the low-resolution basis,
\begin{align}
&\z^\star = \arg\min_{\z} || \x_L - \sum_{j=1}^N (\d_{L,j} \conv \D \z_{j}) ||^2_2 + \beta || \z ||_1
\end{align}
and then convolved with the high-resolution basis to reconstruct the high-resolution image,
\begin{alignat}{4}
&\x_H = \sum_{j=1}^N (\d_{H,j} \conv \z_j)
\end{alignat}

\subsection{Example 6 - Visualizing Object Detection Features}\citep{vondrick_ICCV_2013} presented a method for visualizing HOG features via the inverse mapping,
\begin{align}
&\phi^{-1}(\y) = \arg\min_{\x} || \phi(\x) - \y ||^2_2
\end{align}
where $\x$ is the image to recover, and $\y = \phi(\x)$ is the mapping of the image into HOG space. Direct optimization of this objective is difficult, since it is highly nonlinear through the HOG operator $\phi()$, \ie multiple distinct images can map to the same HOG representation.

One possible approach to approximating this objective is through paired dictionary learning, in a similar manner to \sec{super-resolution}.

Given an image $\x$ and its representation $\y = \phi(\x)$ in the HOG domain, we wish to find two basis sets, $\d_\I$ in the image domain and $\d_\phi$ in the HOG domain, and a common set of sparse reconstruction coefficients $\z$, such that,
\begin{align}
& \x = \sum_{j=1}^N (\d_{\I,j} \conv \z_j) \;\;\;\;\;\;\;\;\;  \y = \sum_{j=1}^N (\d_{\phi,j} \conv \z_j)
\end{align}
Intuitively, the common reconstruction coefficients force the basis sets to represent the same appearance information albeit in different domains. The basis pairs thus provide \emph{a} mapping between the domains.

Optimizing this objective is a straightforward extension of the patch based sparse coding used by \citep{vondrick_ICCV_2013},
\begin{alignat}{4}
& \arg\min_{\mathclap{\d_\I, \d_\phi, \z}} \quad && \sum_{i=1}^M || \x_i - \sum_{j=1}^N (\d_{\I,j} \conv \z_{i,j}) ||^2_2 \nonumber \\
&&& + || \phi(\x_i) - \sum_{j=1}^N (\d_{\phi,j} \conv \z_{i,j}) ||^2_2 \nonumber \\
&&& + \beta \sum_{i=1}^M \sum_{j=1}^M || \z_{i,j} ||_1 \nonumber \\
& \st && || \d_{\I,j} ||^2_2 \le 1 \;\;\; \forall \; j \in 1\dots M \nonumber \\
       &&& || \d_{\phi,j} ||^2_2 \le 1\;\;\; \forall \; j \in 1\dots M
\end{alignat}

Because we are optimizing over entire images rather than independently sampled patches, the bases learned will (i) produce a more unique mapping between pixel features and HOG features (since translations of features are not represented), and (ii) be more expressive for any given basis set size as a direct result of (i).

Image-scale optimization also reduces blocking artefacts in the image reconstructions, leading to more faithful/plausible representations, with potentially finer-grained detail.

\bibliographystyle{plainnat}
\bibliography{citations}

\hrulefill
\begin{appendix}
\section{Log Likelihood Interpretation}
Minimizing the sparse convolutional objective has an equivalent maximum likelihood interpretation,
\begin{alignat}{4}
&\z^\star &&= \arg\max_{\z} \;\; P(\z | \x, \d) \nonumber \\
          &&&= \arg\max_{\x} \;\; P(\x | \d, \z) P(\z)
\end{alignat}
Replacing the product of two probability distributions with the log of their sums, and negating the expression yields,
\begin{alignat}{4}
&\z^\star &&= \arg\min_{\z} \;\; -\log(P(\x | \z, \d)) - P(\z) \;.
\end{alignat}
The corresponding unbiased estimators are,
\begin{alignat}{4}
& -\log(P(\x | \z, \d)) &&\propto \frac{1}{2\sigma^2} || \x - \d \conv \z ||^2_2 \\
& -\log(P(\z)) &&\propto \frac{1}{2b} || \z ||_1
\end{alignat}
After choosing appropriate values for the noise variance,
\begin{alignat}{4}
&\z^\star = \arg\min_{\z} \;\; \half || \x - \d \conv \z ||^2_2 + \beta || \z ||_1 
\end{alignat}

\section{Inversion of a Rank-1 + Scaled Identity Matrix}
\label{sec:inversion_lemma}
The Woodbury matrix identity (matrix inversion lemma) states,
\begin{eqnarray}
(A + UCV)^{-1} = A^{-1} - A^{-1} U ( C^{-1} + V A^{-1} U) ^{-1} V A^{-1}
\end{eqnarray}

Substituting $A = \rho I$, $C = 1$ and $U = x^T$, $V = x$, where $x$ is a rank-1 column matrix gives,
\begin{eqnarray}
(\rho I + xx^T)^{-1} &=& (\rho I)^{-1} - (\rho I)^{-1} x ( 1^{-1} \nonumber \\ && \qquad {} + x^T (\rho I)^{-1} x )^{-1} x^T (\rho I)^{-1} \nonumber \\
			      &=&  (\rho I)^{-1} - (\rho I)^{-1} x ( 1 + \frac{1}{\rho} x^T x )^{-1} x^T (\rho I)^{-1} \nonumber \\
			      &=& (\rho I)^{-1} - (\rho I)^{-1} x ( \rho + x^T x)^{-1} x^T \nonumber \\
			      &=& \frac{1}{\rho} \left( I - \frac{1}{\rho + x^T x} \right ) xx^T
\end{eqnarray}

\end{appendix}
\end{document}